\documentclass[11pt,a4paper]{article}
\usepackage[hyperref]{acl}
\usepackage{times}
\usepackage{latexsym}
\usepackage{booktabs}
\usepackage{graphicx}
\usepackage{xspace}
\usepackage{rotating}
\newcommand\tabrotate[1]{\begin{turn}{90}\rlap{#1}\end{turn}}

\usepackage{amsmath}
\usepackage{booktabs}
\usepackage{adjustbox}
\usepackage{amssymb}
\usepackage{pifont}

\newcommand{\esnli}{\textsc{e-snli}\xspace}
\newcommand{\cose}{\textsc{cos-e}\xspace}

\newcommand{\present}{$\blacksquare$}
\newcommand{\increase}{$\blacktriangle$}
\newcommand{\decrease}{$\blacktriangledown$}
\newcommand{\stable}{$\bullet$}

\newcommand{\doccls}{\textsc{doc. cls.}}

\newcommand{\sa}{\textsc{sentiment analysis}}
\newcommand{\hatespeech}{\textsc{hate speech}}
\newcommand{\re}{\textsc{rel. extraction}}
\newcommand{\mrc}{\textsc{multi choice qa}}
\newcommand{\nli}{\textsc{natural lang. inference}}
\newcommand{\exqa}{\textsc{extractive qa}}

\newcommand{\lr}{\textsc{logreg}}
\newcommand{\nb}{\textsc{naive bayes}}
\newcommand{\svm}{\textsc{svm}}
\newcommand{\cnn}{\textsc{cnn}}
\newcommand{\lstm}{\textsc{lstm}}
\newcommand{\plm}{\textsc{pretr. transformer}}

\newcommand{\highlight}{\textsc{highlight}\xspace}

\newcommand{\freetext}{\textsc{free-text}\xspace}
\newcommand{\semistructured}{\textsc{semi-structured}\xspace}

\newcommand{\reg}{\textsc{regularization}\xspace}
\newcommand{\parsing}{\textsc{data augmentation}\xspace}
\newcommand{\mtl}{\textsc{mtl}\xspace}
\newcommand{\other}{\textsc{other}\xspace}
\newcommand{\irro}{\textsc{i}$\rightarrow$\textsc{ex};\textsc{ex}$\rightarrow$\textsc{o}\xspace}
\newcommand{\ir}{\textsc{i}$\rightarrow$\textsc{ex}\xspace}
\newcommand{\ro}{\textsc{ex}$\rightarrow$\textsc{o}\xspace}

\newcommand{\ood}{\textsc{out-of-domain}\xspace}
\newcommand{\efficiency}{\textsc{efficiency}\xspace}
\newcommand{\bias}{\textsc{bias reduction}\xspace}
\newcommand{\expl}{\textsc{model explanation}\xspace}

\newcommand{\task}{\textsc{task}\xspace}
\newcommand{\model}{\textsc{model}\xspace}
\newcommand{\type}{\textsc{ex. type}\xspace}
\newcommand{\method}{\textsc{method}\xspace}
\newcommand{\eval}{\textsc{results}\xspace}

\usepackage{xcolor,colortbl}

\definecolor{Gray}{gray}{0.94}
\definecolor{LightCyan}{rgb}{0.88,1,1}

\newcolumntype{a}{>{\columncolor{Gray}}c}
\newcolumntype{b}{>{\columncolor{white}}c}

\usepackage{microtype}

\title{A survey on improving NLP models with human explanations}

\author{Mareike Hartmann$^1$  \ \ \ \ \  Daniel Sonntag$^{1,2}$ \\
  $^1$German Research Center for Artificial Intelligence (DFKI), Germany\\
  $^2$Applied Artifical Intelligence (AAI), Oldenburg University, Germany \\
  \texttt{\{mareike.hartmann, daniel.sonntag\}@dfki.de} }

\date{}

\begin{document}
\maketitle
\begin{abstract} 
Training a model with access to human explanations can improve data efficiency and model performance on in- and out-of-domain data. Adding to these empirical findings, similarity with the process of human learning makes learning from explanations a promising way to establish a fruitful human-machine interaction. Several methods have been proposed for improving natural language processing (NLP) models with human explanations, that rely on different explanation types and mechanism for integrating these explanations into the learning process. These methods are rarely compared with each other, making it hard for practitioners to choose the best combination of explanation type and integration mechanism for a specific use-case. 
In this paper, we give an overview of different methods for learning from human explanations, and discuss different factors that can inform the decision of which method to choose for a specific use-case.

\end{abstract}

\section{Introduction}

Training machine learning models with human explanations is considered a promising way for interaction between human and machine that can lead to better models and happier users. If a model is provided with information about \textit{why} a specific prediction should be made for an instance, it can often learn more and faster than if just given the correct label assignment \cite{godbole2004document,zaidan2007using}. This reduces the need for annotated data and makes learning from explanations attractive for use-cases with little annotated data available, for example for adapting models to new domains \cite{yao2021refining} or for personalizing them \cite{kulesza2015principles}. Human explanations also push models to focus on relevant features of the data, preventing them from fitting to spurious correlations in the data \cite{teso2019explanatory}. 
On top of these beneficial effects on model quality, supervision in the form of explanations is in line with human preferences, as users asked to give feedback to a model want to provide richer feedback than just correct labels \cite{stumpf2007toward,amershi2014power,ghai2021explainable}.

\begin{table}[t]
\centering
\resizebox{\linewidth}{!}{
\begin{tabular}{lp{6cm}}
\multicolumn{2}{c}{\esnli}   \\
  Premise & A 2-3 year old \textbf{blond} child is kneeling on a couch.\\
   Hypothesis & The child has \textbf{brown} hair.\\
    Gold label & \texttt{Contradiction}\\
     Free-text & The child would not have brown hair if he/she was blond.\\
     \addlinespace[10pt]
\midrule
 \multicolumn{2}{c}{\cose}   \\
  Questions & What would not be true about a basketball if it \textbf{had a hole} in it but it did not lose its general shape?\\
   Answer options & a) punctured, b) full of air, c) round \\
    Gold label & \texttt{b)} \\
    Free-text & Air cannot stay in any object that has a hole in it.
 \end{tabular}}
 \caption{Examples of highlight (words marked in bold) and free-text explanations in the \esnli dataset \cite{camburu2018snli} for natural language inference and \cose dataset \cite{rajani2019explain} for multiple choice question answering.}\label{t:expls}
 \end{table}

Several approaches for learning from human explanations have been proposed for different tasks (Table \ref{t:modeling}), relying on different types of explanations (Table \ref{t:expls}), and different methods for integrating them into the learning process. In this paper, we review the literature on learning from highlight and free-text explanations for NLP models, listing technical possibilities and identifying and describing factors that can inform the decision for an optimal learning approach that should optimize both model quality and user satisfaction. Our categorization of methods for integrating explanation information (§ \ref{sec:integration}) is similar to the one provided by \citet{hase2021can}.\footnote{Their survey of methods has a broader scope than ours and includes works that improve e.g. image processing models, whereas we exclusively focus on improving NLP models.} Whereas their categorization focuses on contrasting the approaches according to the role of explanation data in the learning process, we focus on how different types of explanations can be integrated with these approaches.

\section{Learning from Explanations}
Highlight and free-text explanations are the most prominent explanation types used to improve NLP models \cite{wiegreffe2021teach}. \textit{Highlight explanations} (\highlight) are subsets of input elements that are deemed relevant for a prediction.\footnote{We follow \citet{wiegreffe2021teach, jacovi2021aligning} in referring to them as highlight explanations.} For text-based NLP tasks, they correspond to sets of words, phrases or sentences. \textit{Free-text explanations} (\freetext) are texts in natural language that are not constrained to be grounded in the input elements and contain implicit or explicit information about why an instance is assigned a specific label. Some recent works rely on \textit{semi-structured text explanations} (\semistructured) \cite{wiegreffe2021teach}, which combine properties of both highlight and free-text explanations. They consist of text in natural language and contain an explicit indication of the input elements that the free-text applies to.\footnote{An overview over NLP datasets with human explanations is provided in \citet{wiegreffe2021teach}.}
 If and how much a model can be improved based on such explanations depends on the amount of information contained in the explanation (§ \ref{sec:infocontent}), and to what extent this information can be integrated into the learning process (§ \ref{sec:integration}). User satisfaction is affected by the effort required to produce explanations and by the difficulty of the task, that might in turn affect explanation quality  (§ \ref{sec:humanfactors}). In the following, we discuss these factors in detail and where possible contrast them with respect to explanation type.

\paragraph{Objectives}
 Approaches for learning from explanations have been evaluated with different objectives in mind, and we introduce the different motivations below and link them with their respective evaluation in Table \ref{t:modeling} (\eval column). Early works for learning from explanations were motivated by making the learning process more \emph{efficient} (\efficiency). Integrating human explanations into the learning process leads to better models trained on the same amount of examples \cite{zaidan2007using}, and to better models trained with annotations collected in the same amount of time \cite{Wang*2020Learning}, i.e. human labor can be used more efficiently. This makes the paradigm useful for use-cases that allow the collection of additional annotations. Information contained in human explanations can make the model generalize better and lead to better predictive performance on \emph{out-of-domain data} (\ood), which is most relevant if the model has to be applied under a distribution shift without access to additional annotations. Even with large amounts of annotated data available, models can fit to noise or unwanted biases in the data \cite{sun-etal-2019-mitigating}, leading to potentially harmful outcomes. Providing human explanations can prevent a model from fitting to such spurious correlations and reduce bias (\bias).\footnote{For this objective, human explanations are often used as feedback in the \emph{explanation-based debugging} setup, where a bug is identified based on a model's explanation for its prediction and fixed by correcting the model explanation \cite{lertvittayakumjorn2021explanation}.} More recently, human explanations have been used in order to improve model explanations (\expl, \citet{strout2019human}) or as targets to enable models to generate explanations in the first place \cite{wiegreffe-etal-2021-measuring}.

\begin{table*}[t]
\centering
\resizebox{\linewidth}{!}{

\begin{tabular}{l ab ab ab ab|b ab ab a|b a b|b ab ab |b ab a}
 
 & \multicolumn{8}{c}{\Large \task} & \multicolumn{6}{c}{\Large \model} & \multicolumn{3}{c}{\Large \type} &  \multicolumn{5}{c}{\Large \method} & \multicolumn{4}{c}{\Large \eval} \\
 \\
 \\
 \\
 \\
 \\
 \\
 \\
 \\
 \\
 \\
\Large Authors & 
\tabrotate{ \textbf{\doccls}} & 
\tabrotate{ \textbf{\sa}} & 
\tabrotate{ \textbf{\hatespeech}} &  
\tabrotate{ \textbf{\re}} & 
\tabrotate{ \textbf{\mrc}} & 
\tabrotate{ \textbf{\nli }} & 
\tabrotate{ \textbf{\exqa}} &
\tabrotate{ \textbf{\other}} &
\tabrotate{ \textbf{\lr}} & 
\tabrotate{ \textbf{\nb}} & 
\tabrotate{ \textbf{\svm}} &  
\tabrotate{ \textbf{\cnn}} & 
\tabrotate{ \textbf{\lstm}} & 
\tabrotate{ \textbf{\plm}} & 
\tabrotate{ \textbf{\highlight}} & 
\tabrotate{ \textbf{\freetext}} &  
\tabrotate{ \textbf{\semistructured}} & 
\tabrotate{ \textbf{\reg}} & 
\tabrotate{ \textbf{\parsing}} & 
\tabrotate{ \textbf{\mtl}} &  
\tabrotate{ \textbf{\irro}} & 
\tabrotate{ \textbf{\other}} &
%
\tabrotate{ \textbf{\ood}} & 
\tabrotate{ \textbf{\efficiency}} &  
\tabrotate{ \textbf{\bias}} & 
\tabrotate{ \textbf{\expl}} \\

\toprule 
 \citet{godbole2004document} & \present & & & & & & & &   
 & &\present & & & &
 \present   & &  &
 & & & & \present &
 & \increase & & \\
 \citet{zaidan2007using} & & \present & & & & & & &  
 & & \present & & & &
 \present   & &  &
 & & & & \present &
 & \increase & & \\
 \citet{zaidan2008modeling}& & \present & & & & & & &  
 & & & & & &
  \present   & &  &
 & & & & \present &
 & \increase & & \\
 \citet{druck2009active}& & & & & & & & \present &  
 & & & & & &  
   \present   & &  & 
 & & & & \present & 
  & \increase & & \\
 \citet{small2011constrained}& & \present & & & & & & &  
 & & \present & & & &
 \present   & &  &
 & & & & \present &
  & \increase& & \\
 \citet{settles2011closing}& \present & & & & & & & &  
 & & & & & & 
 \present   & &  &
 & & & & \present &
  & \increase & & \\
 \citet{kulesza2015principles}& \present & & & & & & & &  
 & \present & & & & & 
 \present   & &  &
 & & & & \present &
  & \increase & & \\
 \citet{zhang2016rationale}& \present & & & & & & & &  
 & & &\present & & &
 \present    & &  &
 & & & \present & &
  & \increase & & \increase \\
 \citet{bao2018deriving}& &\present & & & & & & &  
 & & & & & \present &
 \present   & &  &
 & & & &\present &
  \increase & & & \\
\citet{zhong2019fine}& & \present & & & & & & &   
 & & & & \present & &
 \present   & &  &
 \present & & & & &
  & \increase  & & \stable\\
 \citet{liu-avci-2019-incorporating}& & & \present & & & & & &  
 & & & \present & & &
 \present   & &  &
 \present & & & & &
  &\increase & \increase & \\
  \citet{strout2019human}& & \present&  & & & & & &  
 & & & \present & & &
 \present   & &  &
 \present & & & & &
  &\increase &  & \increase\\
 \citet{rieger2020interpretations}& & \present & & & & & & &  
 & & & & \present & &
 \present   & &  &
 \present & & & & &
  & & \increase & \\
 \citet{stacey2021natural}& & & & & & \present & & &  
 & & & & & \present &
 \present   & &  &
 \present & & & & &
  \increase & \increase & & \\
 \citet{carton2021learn}& \present & & & & \present & \present& & &  
 & & & & & \present &
 \present   & &  &
 & & & \present & &
  & \increase & & \increase \\
 \citet{mathew2021hatexplain}& & & \present & & & & &  & 
 & & & & & &
 \present   & &  &
 \present & & & & &
  & & \increase & \stable \\
 \citet{antognini2021interacting}& & & & & & & & \present & 
 & & & & & $\square$ &
 \present  & &   &
 & & \present & & &
  & \increase & & \increase \\
 \citet{pruthi2020evaluating}& & \present & & & & & \present &  &
 & & & & & \present &
 \present   & &  &
 \present & & \present & & &
  & \increase & & \\
\midrule
\citet{srivastava2017joint}& \present & & & & & & & &  
 \present& \present& & & & &
 &  \present&   &
 & & & & \present &
 & \increase & & \\
\citet{hancock2018training}& & & & \present & & & &  & 
\present & & & & & &
 &  \present &   &
 & \present & & & &
 & \increase & & \\
\citet{Wang*2020Learning}& & \present & & \present & & & &  & 
 & & & & \present & &
 &  \present &  &
 & \present & & & &
 & \increase & & \\
\citet{lee2020lean}& & \present & & \present & & & &  \present &
 & & & &\present & &
 \present  & \present &  &
 & \present & & & &
 & \increase & & \\
\citet{ye2020teaching}& & & & & & & \present &  & 
 & & & & & \present &
 & &  \present  &
 & \present & & & &
 & \increase & & \\
\citet{murty-etal-2020-expbert}& & & & \present & & & &   &
 & & & & & \present &
 &  \present &   &
 & & & & \present &
 & \increase & & \\
\citet{yao2021refining}& & \present & \present & & & & &  & 
 & & & & & \present &
 & &  \present  & 
 & \present & & & &
  \increase & \increase & \increase & \\
\citet{camburu2018snli}& & & & & & \present &  &   &
 & & & &\present & &
 &  \present&  &
 & & \present & \present & &
 \stable & \stable & & \increase \\
\citet{rajani2019explain}& & & & & \present & & &  & 
 & & & & & \present &
 &  \present &   &
 & & & \present & &
 \stable & \increase & & \increase \\
 \citet{kumar2020nile}& & & & & &\present & & &   
 & & & & &\present &
 &  \present&   &
 & & & \present& &
 \decrease & \increase & & \increase \\
 \citet{zhao2021lirex}& & & & & &\present & &  & 
 & & & & &\present &
 \present  & \present&  &
 & & & \present& &
 & &\stable &   \\
 \end{tabular}}
 \caption{An overview over methods for learning NLP tasks from highlight (upper part) and free-text explanations (lower part). The target task (\task), model (\model), explanation type (\type), and integration mechanism (\method) used in the respective work is indicated as \present. $\square$ indicates a transformer model without pre-training. For results reported in the respective paper (\eval), we explicitly mark an observed increase (\increase), decrease (\decrease), or minimal change ($<$1\%, \stable) in the evaluated quantity compared to a baseline without access to explanations.}\label{t:modeling}
 \end{table*}

\subsection{Integrating explanation information}\label{sec:integration}
We now give an overview of different methods\footnote{\citet{hase2021can} derive a framework in which some of these methods can be considered as equal.} that are most commonly applied for integrating the information contained in the human explanation into the model (\method column in Table \ref{t:modeling}). 

   Given an input sequence $\mathbf{x} = (x_1, \cdots, x_L)$ of length $L$, a highlight explanation $\mathbf{a}$ is a sequence of attribution scores $\mathbf{a} = (a_1, \cdots, a_L)$, which is of the same length as $\mathbf{x}$ and assigns an importance of $a_i \in \mathbb{R}$ to input element $x_i$. In practice, $a_i$ is often binary. A free-text explanation $\mathbf{e} = (e_1, \cdots, e_M)$ is a sequence of words of arbitrary length.

\paragraph{Regularizing feature importance}
This is the dominant approach for learning from highlight explanations. The model is trained by minimizing an augmented loss function $\mathcal{L} = \mathcal{L}_\text{CLS} + \mathcal{L}_\text{EXP}$ composed of the standard cross-entropy classification loss $\mathcal{L}_\text{CLS}$ and an additional explanation loss $\mathcal{L}_\text{EXP}$. 
Given a sequence $\mathbf{\hat{a}} = (\hat{a}_1, \cdots, \hat{a}_L)$ of attribution scores extracted from the model, the explanation loss is computed by measuring the distance between gold attributions $a_i$ and model attribution $\hat{a}_i$ according to $\mathcal{L}_\text{EXP}(\mathbf{a}, \mathbf{\hat{a}}) = \sum\limits_i^L \text{dist}(a_i, \hat{a}_i)$. \\
$\mathbf{\hat{a}}$ can be extracted from the model using gradient-based or perturbation-based attribution methods \cite{atanasova-etal-2020-diagnostic}, or attention scores \cite{DBLP:journals/corr/BahdanauCB14}. Intuitively, the model is forced to pay attention to input elements that are highlighted in the highlight explanation. This method is particularly suited for explanation-based debugging, as a user can directly interact with a model by modifying the highlight explanations provided by the model.

\paragraph{Semantic parsing to obtain noisy labels}
This is the dominant approach for learning from free-text explanations. The information contained in the free-text explanations is made accessible via a semantic parser that maps $\mathbf{e}$ to one or more labeling functions $\lambda_i$: $\mathcal{X} \rightarrow \{0,1\}$ \cite{ratner2016data}. $\lambda_i$ is a logical expression executable on input sequence $\mathbf{x}$ and evaluates to 1 if $\mathbf{e}$ applies to $\mathbf{x}$, and 0 otherwise. The set of all labeling functions is then used to assign noisy labels to unlabeled sequences for augmenting the training dataset. Existing methods differ in how the labeling functions are applied to assign noisy labels, e.g. by aggregating scores over multiple outputs or fuzzily matching input sequences. The approach hinges on the availability of a semantic parser, but several works suggest that using a relatively simple to adapt rule-based parser is sufficient for obtaining decent results \cite{hancock2018training}. Table \ref{t:modeling} refers to this approach as \parsing.

\paragraph{Multi-task learning}
In the multi-task learning (MTL) approach \cite{caruana1997multitask}, two models M$_{\text{CLS}}$ and M$_{\text{EXP}}$ are trained simultaneously, one for solving the target task and one for producing explanations, with most of their parameters being shared between them. When learning from highlight explanations, M$_{\text{EXP}}$ is a token-level classifier trained to solve a sequence labeling task to predict the sequence of attributions $\mathbf{a}$. For learning from free-text explanations, M$_{\text{EXP}}$ is a language generation model trained to generate the $\mathbf{e}$.

\paragraph{Explain and predict}
This method was introduced explicitly to improve interpretability of the model, rather than learning from human explanations to improve the target task \cite{lei2016rationalizing}. The idea is to first have the model produce an explanation based on the input instance (\ir), and then predict the output from the explanation alone (\ro), which is meant to assure that the generated explanation is relevant to the model prediction. The approach can be used for both learning from highlight and free-text explanations.\footnote{  \citet{wiegreffe-etal-2021-measuring} provide a recent survey on explain and predict pipelines. For space reasons, the \irro approaches for learning from \highlight explanations listed in their paper are omitted from Table \ref{t:modeling}.} In contrast to the other methods described previously, explain and predict pipelines require explanations at test time. The human explanations are used to train the \ir component, which provides the \ro component with model explanations at test time.

\paragraph{Comparative studies}
We found almost no works that empirically compare approaches for learning from explanations across integration methods or explanation types.
\citet{pruthi2020evaluating} compare \mtl and \reg methods for learning from \highlight explanations. They find that the former method requires more training examples and slightly underperforms regularization. \citet{stacey2021natural} evaluate their \reg method for both \highlight and \freetext explanations. Results are similar for both explanation types, which might be due to the fact that explanations are from the \esnli dataset, where annotators were encouraged to include words contained in the highlight explanation into their free-text explanations.

\subsection{Information content}\label{sec:infocontent}

Besides the choice of method for integrating explanation information, another important factor affecting model performance relates to the information contained in the explanation. Ideally, we could define specific criteria that determine if an explanation is useful for solving a task, and use these criteria for selecting or generating the most beneficial explanations, e.g. as part of annotation guidelines for collecting explanation annotations. 
In the following, we summarize findings of recent works that provide insights for identifying such criteria.
\paragraph{Selecting informative explanations}
Based on experiments with an artificial dataset, \citet{hase2021can} conclude that a model can be improved based on explanations if it can infer relevant latent information better from input instance and explanation combined, than from the input instance alone. This property could be quantified according to the metric suggested by   \citet{pruthi2020evaluating}, who quantify explanation quality as the performance difference between a model trained on input instances  and trained with additional explanation annotations.  \citet{carton2021learn} find that models can profit from those highlight explanations which lead to accurate model predictions if presented to the model in isolation. 
\citet{carton2020evaluating} evaluate human highlight explanations with respect to their comprehensiveness and sufficiency, two metrics usually applied to evaluate the quality of model explanations \cite{yu2019rethinking}, and observe that it is possible to improve model performance with 'insufficient' highlight explanations. In addition, they find that human explanations do not necessarily fulfill these two criteria, indicating that they are not suited for identifying useful human explanations to learn from. As the criteria listed above depend on a machine learning model, they cannot completely disentangle the effects of information content and how easily this content can be accessed by a model. This issue could be alleviated by using model-independent criteria to categorize information content. For example, \citet{aggarwal2021explanations} propose to quantify the information contained in a free-text explanation by calculating the number of distinct words (nouns, verbs, adjectives, and adverbs) per explanation. 

\paragraph{Explanation type}
The works described above focus on identifying informative instances of explanations of a given explanation type. On a broader level, the information that can possibly be contained in an explanation is constrained by its type. Highlight explanations cannot carry information beyond the importance of input elements, e.g. world-knowledge relevant to solve the target task, or causal mechanisms \cite{tan2021diversity}. Hence, free-text explanations are assumed to be more suitable for tasks requiring complex reasoning, such as natural language inference or commonsense question answering \cite{wiegreffe2021teach}. While this assumption intuitively makes sense, it would be useful to more formally characterize the information conveyed in an explanation of a specific type, in order to match it with the requirements of a given target task. \citet{tan2021diversity} define a categorization of explanations that might provide a good starting point for characterizing information content. They group explanations into three categories based on the conveyed information: \emph{Proximal mechanisms} convey how to infer a label from the input, \emph{evidence} conveys relevant tokens in the input (and directly maps to highlight explanations), and \emph{procedure} conveys step-by-step rules and is related to task instructions.
With respect to matching requirements of a given target task, \citet{jansen2016s} describe a procedure for generating gold explanations covering specific knowledge and inference requirements needed to solve the target task of science exam question answering, which might be transferred to other tasks for generating informative explanations.

\subsection{Human factors}\label{sec:humanfactors}
Providing explanations instead of just label annotations requires some overhead from the user, which might negatively affect them. \citet{zaidan2007using} found that providing additional highlight explanations took their annotators twice as long as just providing a label for a document classification task.\footnote{We hypothesize that writing a free-text explanations might take longer than marking highlights for a given task, but could not find any comparison between annotation times for both explanation types.} They also point out the necessity to account for human impatience and sloppiness leading to low-quality explanations.
 \citet{tan2021diversity} list several factors that might limit the use of human-generated explanations, including their incompleteness and subjectivity. Most importantly, they point out that we cannot expect human explanations to be valid even if the human can assign a correct label, as providing an explanation requires deeper knowledge than label assignment. 

\section{Take-Aways}
While many approaches for improving NLP models based on highlight or free-text explanations have been proposed, there is a lack of comparative studies across different explanation types and integration methods that could reveal the most promising setup to proceed with. Initial studies on the relation between explanation properties and effect on model quality suggest that the explanation's information content plays a central role. We see a promising avenue in developing model-independent measures for quantifying information content, which could be used to give annotators detailed instructions on how to generate an informative explanation that can benefit the model, or to filter out invalid explanations that could harm model performance. 

\section*{Acknowledgments}
We thank the reviewers for their insightful comments and suggestions. The research was funded by the XAINES project (BMBF, 01IW20005).

\bibliographystyle{acl_natbib}
\bibliography{anthology,acl2021}

\end{document}